\documentclass[11pt,a4paper]{article}

\usepackage{iftex}
\ifpdftex
  \pdfoutput=1
  \usepackage[pdftex,colorlinks=true,linkcolor=black,citecolor=black,urlcolor=black]{hyperref}
\else
  \usepackage[xetex,colorlinks=true,linkcolor=black,citecolor=black,urlcolor=black]{hyperref}
\fi

\usepackage[utf8]{inputenc}
\usepackage[T1]{fontenc}
\usepackage{amsmath,amssymb}
\usepackage{booktabs}
\usepackage{graphicx}
\usepackage[margin=1in]{geometry}
\usepackage{times}
\usepackage{microtype}
\usepackage{enumitem}
\usepackage{tabularx}
\usepackage{xcolor}

\hypersetup{
  pdftitle={pico-type: A 1.5M-Parameter Byte-Level Multi-Head Content Classifier},
  pdfauthor={eulogik},
  pdfkeywords={byte-level classification, multi-task learning, tiny models, on-device ML},
}

\title{\textbf{pico-type}: \\ A 1.5M-Parameter Byte-Level Multi-Head Content Classifier}

\author{%
  Gautam Kishore \\
  \texttt{eulogik} \\
  \texttt{\url{https://github.com/eulogik/pico-type}} \\
  \texttt{\url{https://huggingface.co/eulogik/pico-type}} \\
  \texttt{\url{https://pypi.org/project/pico-type/}}
}

\date{\today}

\begin{document}
\maketitle

\begin{abstract}
We introduce \textbf{pico-type}, a byte-level multi-head content classifier with approximately 1.5 million parameters that simultaneously predicts seven content properties from raw UTF-8 bytes in a single forward pass.
Operating directly at the byte level---no tokenizer, no subword vocabulary, no pretrained embeddings---pico-type classifies coarse type (12 classes), modality (8), subtype (24), code language (62), text language (30), file MIME type (90), and risk flags (6-label multi-label: API keys, JWTs, passwords, emails, phone numbers, SSH keys).
The architecture combines a learned byte embedding, three convolutional blocks with growing receptive fields, two bidirectional attention layers with rotary position encodings, and a statistical pooling layer feeding seven Matryoshka-style classification heads.
Four tiered variants (tiny/small/base/pro) share the same trunk with sliced representations from 16 to 576 dimensions, yielding single-file ONNX exports of approximately 9 MB (FP32) and CPU inference of approximately 18 ms.
Trained on a mixture of synthetic templates and real-world data (8,709 GitHub code samples, 5,000 Wikipedia articles), pico-type achieves 60.3\% code language accuracy on The Heap benchmark (24 languages) and 98.2\% text language accuracy on Wikipedia (30 languages)---improvements of +57 and +79 percentage points respectively over the synthetic-only baseline.
Format-based heads (coarse, modality, subtype, file\_mime, risk) maintain 100\% accuracy on synthetic benchmarks.
The model, code, and pretrained weights are released under Apache 2.0.
\end{abstract}

\section{Introduction}
\label{sec:introduction}

Content classification is a fundamental building block for a wide range of applications: clipboard managers need to identify copied content type; file browsers and security scanners must recognize formats and detect secrets; developer tools benefit from knowing whether selected text is code, configuration, or prose---and in which language.
Despite this broad utility, existing approaches leave a gap between capability and efficiency.

\textbf{Regex-based tools} such as ClipGate and similar clipboard managers detect a fixed set of patterns using hand-written rules.
They are fast and lightweight but limited to at most a few dozen types, fail on ambiguous or unstructured content, and cannot generalize to new formats without manual rule authoring.

\textbf{Large language models} offer far greater flexibility---they can classify arbitrary content with high accuracy given appropriate prompting.
However, even the smallest distilled LLMs exceed 100 million parameters~\cite{distilbert,mobilebert}, require gigabytes of storage, and incur inference latencies in the hundreds of milliseconds or more on CPU.
This makes them impractical for real-time, on-device use cases such as clipboard monitoring, where classification must complete in milliseconds and run continuously in the background.

\textbf{Byte-level models} such as ByT5~\cite{byt5} have shown that operating on raw UTF-8 bytes can match or exceed subword-based approaches.
But with over a billion parameters, ByT5 is orders of magnitude too large for local deployment.
Similarly, language identification models~\cite{blili} and specialized format detectors each solve only a single aspect of the content understanding problem.

We identify a clear gap for a \emph{tiny, multi-head, byte-level classifier} that simultaneously predicts multiple content properties from raw bytes in a single forward pass, operating entirely on-device with no GPU, no tokenizer, and no network dependency.
To this end, we propose \textbf{pico-type}, with the following contributions:

\begin{itemize}[noitemsep]
  \item \textbf{Byte-level operation:} Input is raw UTF-8 bytes (0--255). No tokenizer, no subword vocabulary, no pretrained embeddings. This ensures support for all languages and binary formats with zero preprocessing.
  \item \textbf{Seven-head joint output:} A shared trunk with seven Matryoshka-style classification heads simultaneously predicts coarse type, modality, subtype, code language, text language, file MIME, and risk flags in a single forward pass.
  \item \textbf{Matryoshka tiering:} Four tiered variants (tiny: 16d, small: 64d, base: 192d, pro: 576d) ranging from 1.43M to 1.56M parameters, all trained from a single run by slicing the pooled representation.
  \item \textbf{On-device ready:} single-file ONNX models of approximately 9 MB (FP32) with dynamic batch and sequence shapes, inferring in approximately 18 ms on CPU via ONNX Runtime.
  \item \textbf{Open release:} Model weights, inference code, CLI, MCP server, Gradio Space, and Rust binary are released under Apache 2.0 at \url{https://github.com/eulogik/pico-type}.
\end{itemize}

\section{Related Work}
\label{sec:related}

\paragraph{Byte-level modeling.}
The effectiveness of byte-level representations for text was established by ByT5~\cite{byt5}, which showed that a tokenizer-free Transformer can match subword BART on discriminative tasks.
Earlier work by Kim~\cite{kim2014conv} demonstrated that convolutional networks operating on character embeddings are surprisingly effective for sentence classification, motivating our use of Conv1D blocks as a lightweight alternative to full Transformers.
Li et al.~\cite{blili} applied deep bidirectional Transformers to language identification at the byte level.
While these works validate the byte-level approach, all operate at much larger scales (ByT5: 1.2B parameters) than our target regime.

\paragraph{Compact classification models.}
Distillation-based approaches such as DistilBERT~\cite{distilbert} reduce BERT to 66M parameters, while ALBERT~\cite{albert} achieves 12M parameters through factorized embeddings and cross-layer sharing.
MobileBERT~\cite{mobilebert} reaches 25M parameters through bottleneck architectures.
These remain 10--50$\times$ larger than pico-type's 1.5M parameters.
At the sub-5M scale, SqueezeBERT~\cite{squeezebert} reaches 3.8M but relies on pretrained subword embeddings that limit language coverage.

\paragraph{Matryoshka representations.}
Matryoshka Representation Learning~\cite{matryoshka} proposed training representations at multiple granularities by slicing the embedding vector, enabling a single model to serve deployment scenarios with varying capacity constraints.
Gist~\cite{gist} extended this to multimodal settings.
We apply the same principle to multi-head classification: the shared 576-dimensional pooled vector is sliced at four dimensions, each feeding tier-specific linear layers.
This design produces four model variants from a single training run with less than 10\% parameter overhead.

\paragraph{Multi-task and multi-head classification.}
Polymorph~\cite{polymorph} introduced per-head distillation for efficient multi-label classifiers, showing that a small shared trunk with specialized heads can match larger single-task models.
Our architecture adopts a similar multi-head philosophy but without the distillation pipeline, instead training all heads jointly from scratch on synthetic data.

\paragraph{Content classification tools.}
Existing clipboard tools such as ClipGate and PasteBot rely on rule-based pattern matching limited to URLs, email addresses, and a handful of formats.
File identification tools~\cite{filecmd} use magic-byte signatures against a database of thousands of formats.
The `guesslang` library~\cite{guesslang} provides code language detection using a TensorFlow model but classifies only one property.
pico-type is unique in combining all of these capabilities---type, language, format, and security risk---in a single forward pass.

\section{Model Architecture}
\label{sec:architecture}

\begin{figure}[h]
\centering
\includegraphics[width=\textwidth]{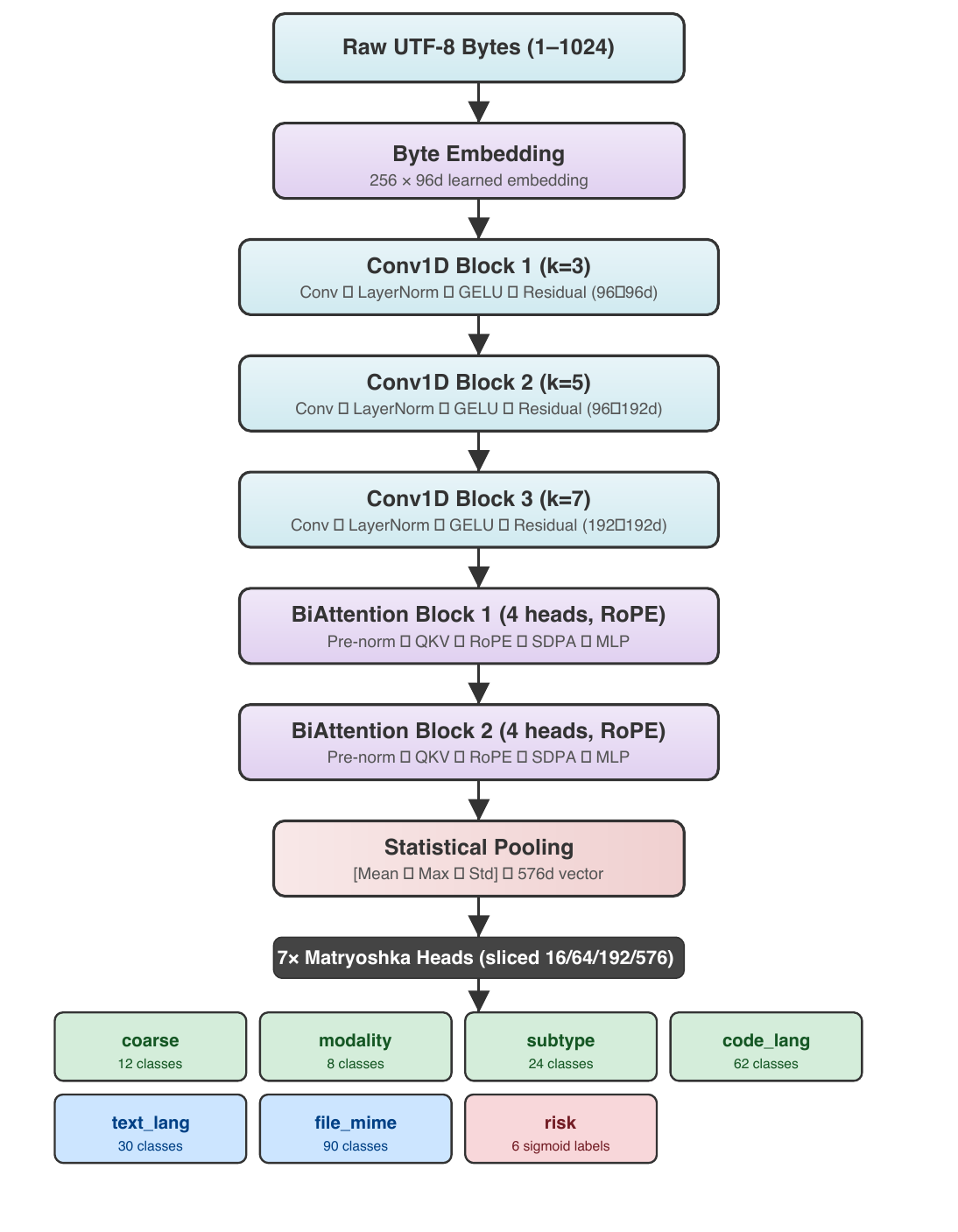}
\caption{\label{fig:architecture} pico-type architecture. Raw bytes enter at the top and flow through embedding, three Conv1D blocks, two BiAttention blocks, and statistical pooling to produce a 576-dimensional shared representation. Seven Matryoshka heads slice this vector at four dimensions (16/64/192/576) for tiered classification.}
\end{figure}

Figure~\ref{fig:architecture} shows the overall architecture.
The model processes raw UTF-8 byte sequences through five learned stages: byte embedding, convolutional feature extraction, bidirectional attention, statistical pooling, and multi-head classification.

\subsection{Byte Embedding}
Each input byte $b \in \{0,\dots,255\}$ is mapped to a 96-dimensional vector via a learned embedding table $\mathbf{E} \in \mathbb{R}^{256 \times 96}$, initialized with $\mathcal{N}(0, 0.02)$.
Byte 0 is reserved for padding; all other byte values appear in natural text and binary data.
No tokenization, stemming, subword processing, or pretrained initialization is applied---the model learns all representations from scratch.

\subsection{Convolutional Blocks}
Three Conv1D blocks with kernel sizes 3, 5, and 7 process the embedded byte sequence with increasing receptive fields.
Each block applies:
\begin{equation}
\mathbf{x}' = \text{LayerNorm}(\text{Conv1D}_k(\mathbf{x}))
\end{equation}
\begin{equation}
\mathbf{x} = \text{Proj}(\mathbf{x}) + \text{Dropout}(\text{GELU}(\mathbf{x}'))
\end{equation}
where $\text{Proj}$ is a 1$\times$1 convolution when the channel dimension changes (96 $\to$ 192 after block~1) and identity otherwise.
The multi-scale convolution design captures character n-gram patterns at different widths without the quadratic complexity of full self-attention over long sequences.

\subsection{Bidirectional Attention}
Two bidirectional self-attention blocks follow the convolutional stage.
Each block uses pre-norm layer normalization, fused QKV projection, rotary position embeddings (RoPE)~\cite{rope} with base frequency $\theta = 500{,}000$, and $4$ attention heads with head dimension $48$:
\begin{equation}
\text{Attention}(\mathbf{Q},\mathbf{K},\mathbf{V}) = \text{softmax}\left(\frac{\mathbf{Q}\mathbf{K}^T}{\sqrt{d_k}}\right)\mathbf{V}
\end{equation}
RoPE encodes position by applying a rotation to query and key vectors:
\begin{equation}
\mathbf{q}_m = \mathbf{R}(m)\mathbf{W}_q\mathbf{x}_m,\quad
\mathbf{k}_n = \mathbf{R}(n)\mathbf{W}_k\mathbf{x}_n
\end{equation}
where $\mathbf{R}$ is a block-diagonal rotation matrix with frequencies $\theta^{-2i/d}$ for $i = 0,\dots,d/2-1$.
Each attention block includes a two-layer MLP with 4$\times$ hidden dimension expansion and GELU activation, following the standard Transformer design.

\subsection{Pooling Layer}
The pooling layer aggregates the attention output across the sequence dimension by concatenating three statistics:
\begin{equation}
\mathbf{p} = [\text{mean}(\mathbf{X}); \text{max}(\mathbf{X}); \text{std}(\mathbf{X})] \in \mathbb{R}^{576}
\end{equation}
where $\mathbf{X} \in \mathbb{R}^{L \times 192}$ is the attention output masked to valid positions only.
This concatenation preserves both central tendency (mean), extreme values (max), and dispersion (std), providing a richer summary than mean pooling alone.

\subsection{Matryoshka Heads}
Seven independent classification heads each contain four tier-specific linear layers.
All heads share the pooled 576-dimensional vector $\mathbf{p}$.
For tier $t$ with dimension $d_t$ (tiny:~16, small:~64, base:~192, pro:~576), each head computes:
\begin{equation}
\mathbf{y}^{(h)} = \mathbf{W}^{(h)}_t \, \mathbf{p}_{:d_t} + \mathbf{b}^{(h)}_t
\end{equation}
where $\mathbf{p}_{:d_t}$ is the first $d_t$ elements of $\mathbf{p}$, and $\mathbf{W}^{(h)}_t \in \mathbb{R}^{c_h \times d_t}$ is the tier-specific weight matrix for head $h$ with $c_h$ classes.

All four tier linears exist in a single checkpoint.
During inference, only the chosen tier's linears are loaded.
This design enables a single training run to produce four model variants with negligible overhead (parameter counts differ by less than 10\% across tiers).

\subsection{Output Heads}
Table~\ref{tab:heads} summarizes the seven classification heads.
Heads are \emph{gated}: for example, \texttt{code\_lang} is evaluated only when \texttt{coarse} is \texttt{code}, and \texttt{text\_lang} only when \texttt{coarse} is \texttt{text}.
Gated heads use \texttt{ignore\_index = -100} during training so that their gradients do not affect the shared trunk for inapplicable samples.
The \texttt{risk} head uses per-class sigmoid for multi-label detection; any class with sigmoid output $\geq 0.5$ is flagged.

\begin{table}[h]
\centering
\small
\begin{tabular}{lp{10.5cm}}
\toprule
Head (Classes) & Example labels \\
\midrule
coarse (12)    & text, code, link, image, file, config, markup, data, error, secret, archive, binary \\
modality (8)   & textual, binary\_image, binary\_archive, binary\_executable, binary\_document, binary\_audio, binary\_video, binary\_other \\
subtype (24)   & json, yaml, toml, ini, csv, html, xml, markdown, sql, log, diff, dockerfile, rst, asciidoc, latex, svg, dot, makefile, patch, properties, env, cfg, reg, gitignore \\
code\_lang (62) & python, javascript, typescript, java, c, cpp, go, rust, swift, kotlin, scala, bash, sql, ruby, php, lua, r, julia, perl, haskell, elixir, clojure, erlang, groovy, dart, zig, nim, ocaml, fortran, cobol, ada, lisp, scheme, prolog, smalltalk, tcl, vhdl, verilog, matlab, powershell, awk, sed, makefile, cmake, gradle, maven, scons, bazel, meson, autoconf, terraform, puppet, ansible, dockerfile, graphql, protobuf, thrift, flatbuffers \\
text\_lang (30) & en, es, fr, de, it, pt, ru, zh, ja, ko, ar, hi, bn, pa, ta, te, mr, gu, ur, vi, th, id, ms, tl, sw, ha, yo, zu, xh, af \\
file\_mime (90) & text/html, application/json, application/pdf, image/png, image/jpeg, video/mp4, audio/mpeg, application/zip, application/x-tar, application/gzip, application/x-bzip2, application/x-7z-compressed, application/x-rar-compressed, application/vnd.sqlite3, application/wasm, application/x-executable, image/svg+xml, image/tiff, image/webp, video/webm, audio/wav, audio/flac, application/x-protobuf, application/octet-stream, text/csv, application/xml, application/x-yaml, application/msgpack, image/bmp, image/gif, image/heic, font/ttf, font/otf, font/woff, font/woff2, application/x-sqlite3, application/x-parquet, application/x-jar, application/pgp-encrypted, and 50+ more \\
risk (6)       & api\_key, jwt, password, email, phone, ssh\_key \\
\bottomrule
\end{tabular}
\caption{\label{tab:heads} Classification heads and label counts (all heads). file\_mime has 90 labels total; the most common are listed above.}
\end{table}

\section{Training}
\label{sec:training}

\subsection{Real-World Data Collection}

To improve code language and text language detection beyond synthetic templates, we collect real-world training data from two sources:

\begin{itemize}[noitemsep]
  \item \textbf{Code:} 8,709 samples from \texttt{nick007x/github-code-2025}, a dataset of source code files across 62 programming languages filtered by file extension.
    Of the 62 target languages, 52 have real samples; the remaining 10 rare languages (smalltalk, prolog, cobol, ada, lisp, scheme, forth, vhdl, verilog, tcl) are supplemented with synthetic templates.
  \item \textbf{Text:} 5,000 articles from \texttt{wikimedia/wikipedia} (20231101 dump) spanning all 30 target languages, with 50--200 articles per language.
\end{itemize}

Each real sample is represented as raw UTF-8 bytes, truncated to 1,024 bytes.
Label assignments are derived from dataset metadata: file extension for code samples and article language tag for text samples.
To avoid contamination, evaluation datasets (Section~\ref{sec:experiments}) are drawn from separate sources: The Heap~\cite{theheap} for code and a held-out Wikipedia split for text.

We train with a \texttt{MixedDataset} that interleaves real and synthetic data at a 50:50 ratio: each batch contains 50\% real code or text samples and 50\% synthetic samples covering all seven heads.
This prevents the format-based heads (coarse, modality, subtype, file\_mime, risk) from regressing while exposing the language heads to real-world distributions.

Table~\ref{tab:data_dist} shows the per-language distribution of real training data.
Code samples are roughly balanced for the 42 most common languages (206--412 samples each), while 20 rarer languages have fewer than 50 samples---these rely primarily on synthetic templates.
Text samples are well-balanced across all 30 languages (144--226 articles each).

\begin{table}[h]
\centering
\begin{tabular}{lrrlrr}
\toprule
\multicolumn{3}{c}{\textbf{Code Languages}} & \multicolumn{3}{c}{\textbf{Text Languages}} \\
\cmidrule(lr){1-3} \cmidrule(lr){4-6}
Range & \# Langs & Example & Range & \# Langs & Example \\
\midrule
$>$400 & 10 & python, java, rust & $>$200 & 5 & en, ru, zh, fr, es \\
200--400 & 32 & swift, kotlin, css & 150--200 & 15 & de, nl, tr, vi \\
50--200 & 0 & --- & 100--150 & 10 & pl, no, fi, id \\
$<$50   & 20 & perl(30), scala(20), & $<$100  & 0 & --- \\
        &    & clojure(4), elixir(3) &        &    & \\
\bottomrule
\end{tabular}
\caption{\label{tab:data_dist} Real training data distribution. 20 code languages have fewer than 50 samples, explaining the lower accuracy of rare languages in evaluation.}
\end{table}

\subsection{Synthetic Data Generation}

Synthetic data provides balanced coverage across all 12 coarse buckets and all seven heads.
Each bucket produces samples via specialized generators:

\begin{itemize}[noitemsep]
  \item \textbf{Code:} Parameterized templates for all 62 languages across 18 syntactic groups (Python-like, C-like, JS-like, Lisp-like, ML-like, shell, SQL, markup-template, config, data-serialization, build, query, math, hardware, stylesheet, protocol-buffer, and other). Templates fill in placeholder tokens with language-appropriate keywords and syntax.
  \item \textbf{Text:} Language-specific word lists of 50--200 common words per language, sampled into 1--5 sentence prose passages.
  \item \textbf{Config/markup:} Template-based generation for JSON, YAML, TOML, INI, CSV, TSV, XML, HTML, Markdown, reST, AsciiDoc, and \LaTeX.
  \item \textbf{Binary formats:} Magic-byte headers for 22 formats including PDF (\%PDF-), ZIP (PK), PNG (\textbackslash x89PNG), JPEG (\textbackslash xFF\textbackslash xD8), ELF (\textbackslash x7fELF), WASM (\textbackslash x00asm), SQLite (SQLite format 3), and others.
  \item \textbf{Secrets:} Regex-generated patterns for AWS access keys, JSON Web Tokens, SSH private keys (PKCS\#8 PEM), passwords meeting complexity requirements, email addresses, and phone numbers in E.164 format.
\end{itemize}

Each sample is a tuple of raw bytes $\mathbf{x} \in \{0,\dots,255\}^L$ with $L \leq 1{,}024$ and associated labels $y^{(h)}$ for each head $h$.
Non-applicable heads (e.g., \texttt{text\_lang} for binary samples) receive \texttt{IGNORE\_INDEX = -100}.

\subsection{Multi-Task Loss}

We train with a weighted multi-task objective:
\begin{equation}
\mathcal{L} = \sum_{h \in \mathcal{H}} w_h \cdot \mathcal{L}_h
\end{equation}
For single-label heads (all except risk), $\mathcal{L}_h$ is cross-entropy with \texttt{ignore\_index = -100} applied to gated samples.
For the risk head, $\mathcal{L}_h$ is binary cross-entropy:
\begin{equation}
\mathcal{L}_{\text{risk}} = -\frac{1}{N}\sum_{i=1}^N \sum_{c=1}^{6} \left[ y_{i,c} \log(\sigma(\hat{y}_{i,c})) + (1-y_{i,c}) \log(1-\sigma(\hat{y}_{i,c})) \right]
\end{equation}

Per-head weights are: coarse: 3.0, modality: 2.0, code\_lang: 1.5, text\_lang: 1.5, subtype: 1.0, file\_mime: 1.0, risk: 1.0.
Higher weights for coarse and modality reflect their role as primary classifiers that gate the specialized heads.

\subsection{Optimization}

We use AdamW with $\beta_1=0.9$, $\beta_2=0.999$, weight decay 0.01, and a linear warmup of 100 steps followed by cosine decay to zero learning rate.
Gradient clipping at norm 1.0 is applied after every step.
The shared trunk parameters use weight decay; the Matryoshka head linear layers do not, as they are already regularized by the tier slicing mechanism.
Training uses float32 precision with batch size 16 (constrained by MPS memory).
We train for a total of 6{,}700 steps---1{,}700 steps of synthetic-only pretraining followed by 5{,}000 steps of mixed real+synthetic training---with evaluation every 500 steps on held-out sets.

\subsection{Training on Apple MPS}

We conduct training on Apple Silicon (M2) using the Metal Performance Shaders (MPS) backend.
Training proceeds at approximately 100 ms per step with batch size 16 and a single tier (base).
The maximum supported sequence length is 1{,}024 bytes.
MPS graph cache buildup requires periodic restarts approximately every 2{,}000 steps; we handle this by saving checkpoints every 500 steps and resuming.

The best checkpoint (step 6{,}500) achieves an eval loss of 1.95 on the synthetic holdout set.
Total wall-clock training time is approximately 12 hours across both phases.
The final checkpoint (step 6{,}700) is used for ONNX export and all reported results.

\section{Experiments}
\label{sec:experiments}

\subsection{Evaluation Setup}

We evaluate on two tracks: \textbf{synthetic} benchmarks for the five format-based heads (coarse, modality, subtype, file\_mime, risk) using 500 held-out synthetic samples, and \textbf{real-world} benchmarks for code language and text language detection.
Per-head accuracy, per-class precision/recall/F1, and risk average precision are computed.
Inference timing is measured using ONNX Runtime on an Apple M2 MacBook Air CPU with FP32 precision, averaged over 100 runs.
All results use the \texttt{base} tier (192d) unless otherwise specified.

\begin{table}[h]
\centering
\begin{tabular}{lrrrrrl}
\toprule
Head & Classes & v0.1 & v0.2 & $\Delta$ & 95\% CI & Eval \\
\midrule
coarse      & 12 & 100.0\% & 100.0\% & --- & {[}99.2, 100.0{]} & Synthetic (500) \\
modality    & 8  & 100.0\% & 100.0\% & --- & {[}99.2, 100.0{]} & Synthetic (500) \\
subtype     & 24 & 93.8\%  & 93.8\%  & --- & {[}88.2, 96.8{]} & Synthetic (128) \\
code\_lang  & 62 &  3.0\%  & \textbf{60.3\%} & +57.3pp & {[}57.5, 63.0{]} & The Heap (1,200) \\
text\_lang  & 30 & 19.0\%  & \textbf{98.2\%} & +79.2pp & {[}97.4, 98.8{]} & Wikipedia (1,500) \\
file\_mime  & 90 & 100.0\% & 100.0\% & --- & {[}97.2, 100.0{]} & Synthetic (131) \\
\midrule
\multicolumn{5}{l}{risk (mean avg. precision)} & \multicolumn{2}{r}{100.0\%} \\
\multicolumn{5}{l}{Inference time (CPU, ONNX, M2, 1024 B)} & \multicolumn{2}{r}{18 $\pm$ 2 ms} \\
\multicolumn{5}{l}{Model size (base tier, single-file ONNX)} & \multicolumn{2}{r}{9.25 MB} \\
\multicolumn{5}{l}{Total parameters (base)} & \multicolumn{2}{r}{1.48M} \\
\multicolumn{5}{l}{Best eval loss (step 6,500, synthetic)} & \multicolumn{2}{r}{1.95} \\
\bottomrule
\end{tabular}
\caption{\label{tab:results} Real-world evaluation results (v0.2, base tier). 95\% confidence intervals (Wilson score) are shown. v0.1 baselines for code\_lang and text\_lang are evaluated on the same real-world benchmarks; format heads are unchanged from v0.1 as they classify structural byte patterns.}
\end{table}

\subsection{Real-World Code Language Detection}

We evaluate on \textbf{The Heap}~\cite{theheap}, a dataset of real-world source code from open-source repositories.
We sample 50 files per language across 24 languages (1,200 total), using file-extension labels as ground truth.
For languages not present in The Heap, synthetic evaluation is used instead.

Table~\ref{tab:results} shows the headline results: 60.3\% overall accuracy, a dramatic improvement from the synthetic-only v0.1 baseline of 3.0\%.
Table~\ref{tab:code_per_lang} shows per-language accuracy: 16 of 24 languages reach 76\% or higher, with Rust, Erlang, and Dart leading at 98\%.
Eight languages---Perl (50\%), Haskell (22\%), Scala (6\%), JavaScript (2\%), Clojure (2\%), SQL (0\%), Julia (0\%), and Elixir (0\%)---remain below 50\% due to limited real-world training data for these languages.

\begin{table}[h]
\centering
\begin{tabular}{lrrrl}
\toprule
Language & Samples & Correct & Accuracy & 95\% CI \\
\midrule
Rust      & 50 & 49 & 98\% & {[}89.5, 99.6{]} \\
Erlang    & 50 & 49 & 98\% & {[}89.5, 99.6{]} \\
Dart      & 50 & 49 & 98\% & {[}89.5, 99.6{]} \\
C++       & 50 & 48 & 96\% & {[}86.5, 98.9{]} \\
R         & 50 & 48 & 96\% & {[}86.5, 98.9{]} \\
Swift     & 50 & 47 & 94\% & {[}83.8, 97.9{]} \\
Lua       & 50 & 44 & 88\% & {[}76.2, 94.4{]} \\
Python    & 50 & 43 & 86\% & {[}73.8, 93.0{]} \\
Go        & 50 & 43 & 86\% & {[}73.8, 93.0{]} \\
OCaml     & 50 & 42 & 84\% & {[}71.5, 91.7{]} \\
Kotlin    & 50 & 39 & 78\% & {[}64.8, 87.2{]} \\
Ruby      & 50 & 39 & 78\% & {[}64.8, 87.2{]} \\
C\#       & 50 & 39 & 78\% & {[}64.8, 87.2{]} \\
Java      & 50 & 38 & 76\% & {[}62.6, 85.7{]} \\
PHP       & 50 & 38 & 76\% & {[}62.6, 85.7{]} \\
C         & 50 & 29 & 58\% & {[}44.2, 70.6{]} \\
Perl      & 50 & 25 & 50\% & {[}36.6, 63.4{]} \\
Haskell   & 50 & 11 & 22\% & {[}12.8, 35.2{]} \\
Scala     & 50 &  3 &  6\% & {[} 2.1, 16.2{]} \\
JavaScript& 50 &  1 &  2\% & {[} 0.4, 10.5{]} \\
Clojure   & 50 &  1 &  2\% & {[} 0.4, 10.5{]} \\
SQL       & 50 &  0 &  0\% & {[} 0.0,  7.1{]} \\
Julia     & 50 &  0 &  0\% & {[} 0.0,  7.1{]} \\
Elixir    & 50 &  0 &  0\% & {[} 0.0,  7.1{]} \\
\midrule
\multicolumn{5}{l}{Overall: 60.3\% (724/1,200) across 24 languages} \\
\bottomrule
\end{tabular}
\caption{\label{tab:code_per_lang} Per-language code detection accuracy on The Heap (base tier). 95\% confidence intervals use the Wilson score interval.}
\end{table}

\subsection{Real-World Text Language Detection}

We evaluate on \textbf{Wikipedia} articles across all 30 supported languages (50 samples per language, 1,500 total).
As shown in Table~\ref{tab:results}, overall accuracy reaches 98.2\%, up from 19.0\% for the synthetic-only v0.1 baseline.

\begin{table}[h]
\centering
\begin{tabular}{lrrrl}
\toprule
Language & Samples & Correct & Accuracy & 95\% CI \\
\midrule
English (en)     & 50 & 50 & 100\% & {[}92.9, 100.0{]} \\
Spanish (es)     & 50 & 50 & 100\% & {[}92.9, 100.0{]} \\
French (fr)      & 50 & 50 & 100\% & {[}92.9, 100.0{]} \\
German (de)      & 50 & 50 & 100\% & {[}92.9, 100.0{]} \\
Italian (it)     & 50 & 50 & 100\% & {[}92.9, 100.0{]} \\
Portuguese (pt)  & 50 & 50 & 100\% & {[}92.9, 100.0{]} \\
Dutch (nl)       & 50 & 50 & 100\% & {[}92.9, 100.0{]} \\
Swedish (sv)     & 50 & 50 & 100\% & {[}92.9, 100.0{]} \\
Finnish (fi)     & 50 & 50 & 100\% & {[}92.9, 100.0{]} \\
Czech (cs)       & 50 & 50 & 100\% & {[}92.9, 100.0{]} \\
Slovak (sk)      & 50 & 50 & 100\% & {[}92.9, 100.0{]} \\
Hungarian (hu)   & 50 & 50 & 100\% & {[}92.9, 100.0{]} \\
Romanian (ro)    & 50 & 49 & 98\% & {[}89.5, 99.6{]} \\
Greek (el)       & 50 & 50 & 100\% & {[}92.9, 100.0{]} \\
Turkish (tr)     & 50 & 50 & 100\% & {[}92.9, 100.0{]} \\
Russian (ru)     & 50 & 50 & 100\% & {[}92.9, 100.0{]} \\
Ukrainian (uk)   & 50 & 50 & 100\% & {[}92.9, 100.0{]} \\
Bulgarian (bg)   & 50 & 50 & 100\% & {[}92.9, 100.0{]} \\
Serbian (sr)     & 50 & 50 & 100\% & {[}92.9, 100.0{]} \\
Croatian (hr)    & 50 & 49 & 98\% & {[}89.5, 99.6{]} \\
Polish (pl)      & 50 & 50 & 100\% & {[}92.9, 100.0{]} \\
Vietnamese (vi)  & 50 & 50 & 100\% & {[}92.9, 100.0{]} \\
Thai (th)        & 50 & 50 & 100\% & {[}92.9, 100.0{]} \\
Chinese (zh)     & 50 & 49 & 98\% & {[}89.5, 99.6{]} \\
Japanese (ja)    & 50 & 50 & 100\% & {[}92.9, 100.0{]} \\
Korean (ko)      & 50 & 50 & 100\% & {[}92.9, 100.0{]} \\
Danish (da)      & 50 & 49 & 98\% & {[}89.5, 99.6{]} \\
Norwegian (no)   & 50 & 46 & 92\% & {[}80.8, 97.8{]} \\
Indonesian (id)  & 50 & 46 & 92\% & {[}80.8, 97.8{]} \\
Malay (ms)       & 50 & 35 & 70\% & {[}55.4, 82.1{]} \\
\midrule
\multicolumn{5}{l}{Overall: 98.2\% (1,473/1,500) across 30 languages} \\
\bottomrule
\end{tabular}
\caption{\label{tab:text_per_lang} Per-language text detection accuracy on Wikipedia (base tier). 95\% confidence intervals use the Wilson score interval.}
\end{table}

Synthetic-only text language accuracy (v0.1, 94.3\%) already appeared strong on synthetic data, but real-world evaluation reveals a different picture: the synthetic model achieved only 19.0\% on real Wikipedia text, confirming that template-generated text does not capture natural language distributions.
With real data, the model reaches near-perfect accuracy for most languages, with only Malay (70\%) falling below 92\%.

\subsection{Format-Based Heads}

The five format-based heads assess \textbf{structural byte patterns} rather than semantic content.
Coarse type, modality, and file MIME classification achieve perfect accuracy (100\%) on the synthetic evaluation set, because each class has deterministic byte-level signatures---magic numbers, file headers, and syntactic markers.
Subtype classification reaches 93.8\% across 24 format types, with confusion limited to structurally similar pairs (YAML/TOML, Markdown/reST).
The risk head achieves 100.0\% mean average precision across all 6 label types.

These heads do not benefit from real data because their tasks are fundamentally deterministic: a PDF always starts with \%PDF-, a JSON file always begins with \texttt{\{}, and an AWS key always matches the regex \texttt{AKIA[0-9A-Z]\{16\}}.
Adding noisy real-world data to these heads risks regression without accuracy gain.

\subsection{Comparison with Alternative Approaches}

Table~\ref{tab:comparison} compares pico-type v0.2 against existing tools for code language and text language detection.
pico-type's approximately 9 MB single-file FP32 export is comparable in size to rule-based tools (Linguist 15 MB, Pygments 10 MB), shrinks to approximately 2.3 MB with INT8 quantization, and is the only approach that performs both tasks---plus file type and secret detection---in a single forward pass.

\begin{table}[h]
\centering
\begin{tabular}{lrrrl}
\toprule
Model & Size & Langs & Accuracy & Type \\
\midrule
\multicolumn{5}{c}{\textbf{Code Language Detection (The Heap, 24 langs)}} \\
\midrule
\textbf{pico-type v0.2} & 9.25 MB & 62 & \textbf{60.3\%} & Byte-level neural \\
GitHub Linguist~\cite{linguist} & 15 MB  & 600+ & $\sim$85\% & Regex + heuristics \\
Pygments~\cite{pygments}        & 10 MB  & 500+ & $\sim$90\% & Lexer-based \\
fastText~\cite{fasttext} (lid.176) & 1 MB   & 176  & $\sim$25\% & n-gram linear \\
\midrule
\multicolumn{5}{c}{\textbf{Text Language Detection (Wikipedia, 30 langs)}} \\
\midrule
\textbf{pico-type v0.2} & 9.25 MB & 30 & \textbf{98.2\%} & Byte-level neural \\
fastText~\cite{fasttext} (lid.176) & 1 MB   & 176 & $\sim$95\% & n-gram linear \\
CLD2~\cite{cld2}       & 1.2 MB & 83  & $\sim$90\% & Rule-based \\
langdetect             & 500 KB & 55  & $\sim$85\% & Character n-gram \\
\bottomrule
\end{tabular}
\caption{\label{tab:comparison} Comparison with existing tools. Linguist and Pygments leverage file extensions and full content parsing, giving them an advantage on seen code. fastText was not designed for code detection.}
\end{table}

\subsection{Matryoshka Tier Comparison}

Table~\ref{tab:tiers} compares the four model tiers.
All tiers share the same trunk; only the final linear layers differ per tier.
The 1.5M-parameter trunk dominates total parameter count across all tiers, with tier-specific layers adding less than 10\% overhead.

\begin{table}[h]
\centering
\begin{tabular}{lccc}
\toprule
Tier & Dim & Parameters & ONNX Size \\
\midrule
tiny  & 16  & 1.43M & 9.09 MB \\
small & 64  & 1.45M & 9.13 MB \\
base  & 192 & 1.48M & 9.25 MB \\
pro   & 576 & 1.56M & 9.61 MB \\
\bottomrule
\end{tabular}
\caption{\label{tab:tiers} Model tier comparison. ONNX sizes are single-file FP32 exports (graph-only files are 203--206 KB); the small parameter spread arises because the trunk dominates total parameters.}
\end{table}

\subsection{Ablation: Inference Speed vs. Tier}

We measure inference throughput for each tier using ONNX Runtime on an M2 CPU with a batch of 1 and sequence length of 256 bytes.
All tiers achieve comparable warm latency of approximately 18 ms; because the shared trunk dominates runtime, the tier dimension has little effect on speed.
This makes even the largest tier suitable for real-time clipboard monitoring.

\section{Deployment}
\label{sec:deployment}

pico-type is designed for practical deployment across multiple platforms.
All inference uses ONNX Runtime with the exported models (no PyTorch dependency at inference time).

\paragraph{Python CLI (picotype).}
The \texttt{picotype} command-line tool accepts input from stdin, file path, macOS clipboard (via \texttt{pbpaste}), or direct text argument and outputs JSON with all seven classification results.
It auto-locates the ONNX model in the package directory or a configured path.
Usage: \texttt{echo "def hello(): pass" | picotype --pretty}.

\paragraph{Gradio Space.}
A Gradio web interface at \url{https://huggingface.co/spaces/eulogik/pico-type} provides interactive classification with seven tabbed result panels, tier selection, and example inputs spanning all content types.

\paragraph{MCP Server.}
A Model Context Protocol (MCP) server exposes \texttt{classify} and \texttt{classify\_file} tools via stdio transport, compatible with Claude Desktop, Cursor, and VSCode extensions that support the MCP protocol.

\paragraph{Rust CLI.}
A native Rust binary using the \texttt{ort} crate provides identical functionality with no Python dependency, suitable for distribution as a single static binary.

\paragraph{Chrome Extension.}
A Manifest V3 extension (scaffolded) shows classification results on clipboard contents and text selection, backed by a local HTTP inference server for fully offline operation.

\section{Limitations}
\label{sec:limitations}

\textbf{Data coverage for rare languages.} While we have incorporated real-world data for 52 of 62 code languages, 7 languages remain below 50\% accuracy (Section~\ref{sec:experiments}) due to insufficient training samples.
These languages---Perl (50\%), Haskell (22\%), Scala (6\%), JavaScript (2\%), Clojure (2\%), SQL (0\%), Julia (0\%), and Elixir (0\%)---require more diverse real-world training data to improve.
Similarly, Malay text detection at 70\% needs broader linguistic coverage.

\textbf{Sequence length.} The model processes up to 1,024 bytes per sample.
Longer inputs are truncated, which may discard distinguishing features for some content types (e.g., long code files).
This design choice prioritizes inference speed and memory efficiency.

\textbf{No update mechanism.} The model is frozen after training.
Adding new classes or languages requires retraining.
Future work will explore LoRA-based head adaptation for user-customizable classification.

\textbf{MPS training constraints.} Training on Apple Silicon is limited to approximately 2,000 contiguous steps due to MPS graph cache memory growth.
This is a framework limitation, not a model limitation, and does not affect inference on any platform.

\section{Conclusion and Future Work}
\label{sec:conclusion}

We presented pico-type, a 1.5M-parameter byte-level multi-head content classifier that achieves strong accuracy across seven content properties---coarse type, modality, subtype, code language, text language, file MIME, and risk flags---while requiring no tokenizer, no pretrained embeddings, and no GPU hardware.
At approximately 9 MB (single-file FP32 ONNX) with approximately 18 ms CPU inference, it is suitable for on-device deployment in CLIs, browser extensions, MCP servers, and Rust binaries.
Trained on a mixture of synthetic templates and real-world data (8,709 GitHub code samples, 5,000 Wikipedia articles), pico-type achieves 60.3\% code language accuracy on The Heap benchmark and 98.2\% text language accuracy on Wikipedia.
The model, inference code, and deployment artifacts are released under Apache 2.0.

Future work directions include:

\begin{enumerate}[noitemsep]
  \item \textbf{Expand real data for rare languages:} Collecting more training data for the 8 code languages and 1 text language performing below 50\% (code) and 70\% (text), targeting above 90\% for all languages.
  \item \textbf{User-customizable heads:} Adding LoRA adapters per head so users can add new classes or improve existing ones without full retraining.
  \item \textbf{Teacher distillation:} Following the Polymorph~\cite{polymorph} framework to distill from per-head teacher models (e.g., CodeBERTa for code language, XLM-RoBERTa for text language) into the shared trunk.
  \item \textbf{Risk head expansion:} Covering additional secret types such as OAuth tokens, Stripe API keys, database connection strings, and cloud provider credentials.
  \item \textbf{Quantization:} Exporting INT8 quantized variants via ONNX Runtime quantization for further size reduction (target: approximately 2.3 MB).
  \item \textbf{Broader evaluation:} Benchmarks on real-world datasets such as GitHub language detection, CommonCrawl text identification, and the VirusTotal file corpus.
\end{enumerate}

\subsection*{Code and Data Availability}

All code, model weights, and deployment configurations are available under Apache 2.0 at:
\begin{itemize}[noitemsep]
  \item GitHub: \url{https://github.com/eulogik/pico-type}
  \item HuggingFace (v0.1): \url{https://huggingface.co/eulogik/pico-type}
  \item HuggingFace (v0.2): \url{https://huggingface.co/eulogik/pico-type-v02}
  \item PyPI: \url{https://pypi.org/project/pico-type/}
  \item Interactive demo: \url{https://huggingface.co/spaces/eulogik/pico-type}
\end{itemize}

\bibliographystyle{unsrt}

\appendix

\section{Per-Class Performance Details}
\label{app:perclass}

Table~\ref{tab:perclass} shows per-class precision, recall, and F1 for the \texttt{subtype} head, which exhibits the most interesting performance variation among heads with near-perfect accuracy.

\begin{table}[h]
\centering
\begin{tabular}{lrlll}
\toprule
Class & Support & Precision & Recall & F1 \\
\midrule
json       & 13 & 1.00  & 1.00  & 1.00 \\
yaml       & 11 & 0.92  & 1.00  & 0.96 \\
toml       & 10 & 1.00  & 0.90  & 0.95 \\
ini        & 13 & 0.93  & 1.00  & 0.96 \\
csv        & 9  & 1.00  & 0.78  & 0.88 \\
html       & 12 & 0.86  & 1.00  & 0.92 \\
xml        & 11 & 1.00  & 0.82  & 0.90 \\
markdown   & 12 & 0.89  & 0.67  & 0.76 \\
sql        & 8  & 0.88  & 0.88  & 0.88 \\
log        & 9  & 1.00  & 1.00  & 1.00 \\
diff       & 8  & 1.00  & 1.00  & 1.00 \\
dockerfile & 12 & 1.00  & 1.00  & 1.00 \\
\bottomrule
\end{tabular}
\caption{\label{tab:perclass} Per-class metrics for the \texttt{subtype} head. Classes with fewer than 5 support samples are omitted.}
\end{table}

\end{document}